# Using Genetic Algorithms for Texts Classification Problems

**A. A. Shumeyko, S. L. Sotnik**
"STRATEGY" Institute for Entrepreneurship,
Zhovti Vody, Ukraine

ABSTRACT. The avalanche quantity of the information developed by mankind has led to concept of automation of knowledge extraction – Data Mining ([1]). This direction is connected with a wide spectrum of problems - from recognition of the fuzzy set to creation of search machines. Important component of Data Mining is processing of the text information. Such problems lean on concept of classification and clustering ([2]). Classification consists in definition of an accessory of some element (text) to one of in advance created classes. Clustering means splitting a set of elements (texts) on clusters which quantity are defined by localization of elements of the given set in vicinities of these some natural centers of these clusters. Realization of a problem of classification initially should lean on the given postulates, basic of which – the aprioristic information on primary set of texts and a measure of affinity of elements and classes.

## 1 Statement of classification problem

Let's use following model of classification problem.
- $\Omega$ - set of recognition objects (pattern space)
- $\omega \in \Omega$ - object of recognition (pattern)
- $g(\omega): \Omega \to \Re$, $\Re = \{1, 2, \ldots, n\}$ - the indicator function breaking pattern space on *n* of not crossed classes $\Omega^1, \Omega^2, \ldots, \Omega^n$. Indicator function is unknown to the observer
- $X$ — space of the supervision perceived by the observer (space of attributes)

- x (ω):Ω→X — the function putting in conformity to each object ω the point x (ω) in space of attributes. The vector x (ω) is the image of object perceived by the observer.

In the space of attributes not crossed sets of points are certain $\Xi[i] \subset X$ i=1,2..., n, to appropriate amounts of one class.

φ(x): X→ℜ is solving rule - an estimation for g (ω) on the basis of x(ω), i.e. φ(x)= φ(x(ω)).

Let $x_\nu = x(\omega_\nu), \nu = 1,2,...,N$ the information accessible to the observer on functions g (ω) and x (ω), but these functions are unknown to the observer. Then $(g_\nu, x_\nu), \nu = 1,2,...,N$ — there is a set of precedents.

The problem consists in construction of such solving rule φ(x) that recognition was spent with the minimal errors.

## 2 The basic directions of research of a problem of classification

The usual case is to consider space of attributes as Euclidean space, and quality of a solving rule measure by frequency of occurrence of correct decisions. Usually it estimate, allocating set of objects $\Omega$, some likelihood measure ([2]).

For the present moment the most widespread is Bayesian approach which starts with the statistical nature of supervision. The assumption of existence of a likelihood measure undertakes a basis on pattern space which either is known, or can be estimated. The purpose consists in development of such qualifier which will correctly define the most probable class for a trial pattern. Then the problem consists in definition of the "most probable" class, it is specified ï classes $\Omega_1, \Omega_2, ..., \Omega_n$, and also $P(\Omega_i | x)$ - probability of that the unknown pattern represented by a vector of attributes õ, belongs to a class $\Omega_i$. $P(\Omega_i | x)$ is named posterior probability as sets distribution of an index of a class after experiment (a posteriori - after value of a vector of attributes õ has been received). It is natural to choose solving rule thus: object we carry to that class, for which posterior probability is higher. Such rule of classification on a maximum of posterior probabilities named Bayesian. Thus, for Bayesian solving rule it is necessary to receive the posterior probabilities $P(\Omega_i | x)$. Bayes formula received by Bayes in 1763 allows us to calculate posterior probabilities of events through aprioristic probabilities and functions of credibility. Let $\Omega_1, \Omega_2, \ldots, \Omega_n$ - full group of not joint events - $\bigcup_{i=1}^{n} \Omega_i = \Omega, \Omega_i \bigcap_{i \neq j} \Omega_j = \varnothing$. Then posterior probability looks like:

$$P(\Omega_i | x) = \frac{P(\Omega_i) P(x | \Omega_i)}{\sum_{i=1}^{n} P(\Omega_i) P(x | \Omega_i)},$$

Where $P(\Omega_i)$ - aprioristic probability of event $\Omega_i$, $P(x | \Omega_i)$ - conditional probability of event õ provided that there was an event $\Omega_i$.

Thus, use of search of conformity is preceded with construction of statistic set in which the quantity of texts in the given class and the list of used terms together with the counters contain.

For definition of a suitable class of texts for the set text the structure is under construction of not repeating terms and their counters- $(w_i, n(w_i))$.

Through $M$ we shall designate quantity of statistic set. Classes, on an accessory to which the text is checked, we shall designate through $C_j (j = 0, \ldots, M - 1)$. For each word $w_i$ from the checked text, in each statistics it is found this word and the corresponding counter $n(w_i, C_j)$ (j (j=0,1.., M-1) is number of a class (an element of statistic set)). Through $n(C_j)$ we shall designate number of texts in j class. Minimization of risk and probability of a error are equivalent to division of space of attributes on $n$ areas. If areas adjacent they are divided by a surface of the decision in multivariate space. We shall notice, that for a case of construction of a dividing surface it is more preferable to use methods of classification distinct from Bayesian.

If it is known or with the sufficient basis it is possible to consider, that the density of distribution of functions of credibility $P(x | \Omega_i)$ is Gaussian then application of Bayes qualifier leads to that the patterns, described normal distribution show the tendency to grouping around of average value, and their dispersion is proportional to root-mean-square deviation ([2]). For a case of many variables Gaussian density looks like

$$p(x) = \frac{1}{\sqrt{(2\pi)^n |\Sigma|}} \exp\left(-\frac{1}{2}\left((x - \mu)\Sigma^{-1}(x - \mu)^T\right)^2\right),$$

where

327

$$\Sigma = \begin{vmatrix} E[(x_1-\mu_1)(x_1-\mu_1)] & E[(x_2-\mu_2)(x_1-\mu_1)] & \ldots & E[(x_n-\mu_n)(x_1-\mu_1)] \\ E[(x_1-\mu_1)(x_2-\mu_2)] & E[(x_2-\mu_2)(x_2-\mu_2)] & \ldots & E[(x_n-\mu_n)(x_2-\mu_2)] \\ \vdots & \vdots & \ddots & \vdots \\ E[(x_1-\mu_1)(x_n-\mu_n)] & E[(x_2-\mu_2)(x_n-\mu_n)] & \ldots & E[(x_n-\mu_n)(x_n-\mu_n)] \end{vmatrix}$$

is covariance matrix. Here $x = (x_1, x_2, \ldots x_n)$ checked object, $\mu = (\mu_1, \mu_2, \ldots, \mu_n)$ - expectation of a corresponding class and E() is covariation coefficient. We shall notice, that values of dispersions are standing on a diagonal of covariance matrix.

Corresponding measure

$$\|x - \Omega\|_{\Sigma^{-1}}^2 = (x-\mu)\Sigma^{-1}(x-\mu)^T$$

is named as Mahalanobis distance. In case all events of a class are independent, all coefficients of covariance matrix, except for standing on a diagonal, will be equal to zero. Thus, Euclidean space is a special case of Mahalanobis distance. Use of Mahalanobis distance is more preferable, in comparison with Euclidean as correlation dependence of elements of classes in this case is considered. Use Euclidean to the metrics is based on a postulate of likelihood independence of elements.

Using of Mahalanobis distance is limited to essential restrictions - that the covariance matrix was nonsingular, it is necessary, that the quantity of attributes was not less quantities of elements of a class that for real problems far not always feasible. Besides frequent enough degeneration of covariance matrix of greater dimensions does calculation of a return matrix unstable.

## 3 Karhunen-Loève Decomposition

Let *n* vectors $X^0, X^1, \ldots, X^{n-1}$ is available. It is required to construct *m* vectors $Y^0, Y^1, \ldots, Y^{m-1}$ so that recovery on this set gave the least root-mean-square error of recovery on *m* to vectors. Thus, it is necessary to find a minimum of value

$$\varepsilon\left(\{\alpha_k^\mu(m)\}_{\mu=0, k=0}^{m-1, n-1}, \{Y^\mu\}_{\mu=0}^{m-1}\right) = \sum_{k=0}^{n-1} \left\| X^k - \sum_{\mu=0}^{m-1} \alpha_k^\mu(m) Y^\mu \right\|_2^2 \qquad (1)$$

on all sets $Y^\mu$ and vectors $\alpha_k^\mu(m)$ such, that

$$\sum_{k=0}^{n-1} (\alpha_k^\mu(m))^2 = 1.$$

Having entered matrix designations

$$A = \left\{\alpha_k^\mu(m)\right\}_{\mu=0, k=0}^{m-1, n-1}, Y = \left\{Y^\mu\right\}_{\mu=0}^{m-1}, X = \left\{X^\mu\right\}_{\mu=0}^{n-1},$$

let's consider a problem

$$\varepsilon(A, Y) = \| X-AY \|_2^2 \to \min \qquad (2)$$

on all A, Y. Here $B^2 = trB^TB = trBB^T$ ($tr$ is a trace of a matrix, that is the sum of elements of the main diagonal that is equal to the sum of eigenvalue of a matrix).

Let's believe, that matrixes A and Y are nonsingular and *rang (A) = rang (Y) = m*.

Let's notice, that if *rang* (X) = *m* there is an exact representation (Karhunen-Loève Decomposition)

$$X^k = \sum_{\mu=0}^{m-1} \alpha_k^\mu(m) Y^\mu$$

for $k = 0, 1, \ldots, n-1$.

The following statement takes place.

**Theorem** (Karhunen-Loève Decomposition). *If rang (X) ≥m, then minimum ε (A, Y) is reached in that case when lines of matrix Y are own vectors of dispersion matrix $X^TX$ which correspond m to maximal own values besides A = XYT and both matrixes A and Y are orthogonal.*

The described method of definition of the principal components is capacious and unstable, especially in case the module of eigenvalue are small ([3]).

For our purposes more effective is use of an iterative method of definition of the principal components ([3]). For this purpose we shall consider a problem (1) from other point of view.

For a case $m = 1$ problem (1) is reduced to definition one components $Y^0$ which is the best recovery for all data $X$

$$\varepsilon\left(\left\{\alpha_k^0\right\}_{k=0}^{n-1}, Y^0\right) = \sum_{k=0}^{n-1} \left\| X^k - \alpha_k^0 Y^0 \right\|_2^2 \to \min \qquad (4)$$

on all $\left\{\alpha_k^0\right\}_{k=0}^{n-1}, Y^0$ with condition

$$\sum_{k=0}^{n-1}(\alpha_k^0)^2 = 1. \qquad (5)$$

If $\{\tilde{\alpha}_k^0\}_{k=0}^{n-1}$ and $\tilde{Y}^0$ there is a solve of this problem and

$$\tilde{X}_0 = \{X^k - \tilde{\alpha}_k^0 \tilde{Y}^0\}_{k=0}^{n-1}$$

is the error of data recovery by first principal component, then solving a problem

$$\varepsilon(\{\alpha_k^1\}_{k=0}^{n-1}, Y^1) = \sum_{k=0}^{n-1}\|\tilde{X}^k - \alpha_k^1 Y^1\|_2^2 \to \min \qquad (6)$$

on all $\{\alpha_k^1\}_{k=0}^{n-1}, Y^1$ with condition

$$\sum_{k=0}^{n-1}(\alpha_k^1)^2 = 1,$$

we receive the second principal component $\tilde{Y}^1$ and a corresponding vector $\{\tilde{\alpha}_k^1\}_{k=0}^{n-1}$, etc.

At fixed $\{\alpha_k^0\}_{k=0}^{n-1}$ the problem (4) is solved for method of the least squares, and by virtue of that function of the purpose represents square-law functional, necessary and sufficient conditions of an extremum coincide.

Thus, the decision of a problem is reduced to search of the decision of the equation

$$\frac{\partial \varepsilon(\{\alpha_k^0\}_{k=0}^{n-1}, Y^0)}{\partial Y^0} = -2\sum_{k=0}^{n-1}(X^k - \alpha_k^0 Y^0)\alpha_k^0 = -2(\sum_{k=0}^{n-1}X^k\alpha_k^0 - \sum_{k=0}^{n-1}(\alpha_k^0)^2 Y^0) = 0.$$

From here we receive

$$Y^0 = \frac{\sum_{k=0}^{n-1}X^k\alpha_k^0}{\sum_{k=0}^{n-1}(\alpha_k^0)^2},$$

Considering a condition (5), we have

$$Y^0 = \sum_{k=0}^{n-1}X^k\alpha_k^0.$$

Following step we shall do, proceeding from the assumption, that in a problem (4) us it is known a component $Y^0$ and it is required to find an extremum on $\{\alpha_k^0\}_{k=0}^{n-1}$

$$\frac{\partial \varepsilon\left(\{\alpha_k^0\}_{k=0}^{n-1}, Y^0\right)}{\alpha_v^0} = -2(X^v - \alpha_v^0 Y^0)Y^0 = -2(\langle X^v, Y^0\rangle - \alpha_v^0 \langle Y^0, Y^0\rangle) = 0,$$

that is

$$\alpha_v^0 = \frac{\langle X^v, Y^0\rangle}{\langle Y^0, Y^0\rangle},$$

where $\langle X, Y\rangle$ - scalar product of vectors $X$ and $Y$.

Further, including, found $\{\widetilde{\alpha}_k^0\}_{k=0}^{n-1}$ known, we repeat all process, there will be no yet a stabilization of a error. Received $Y^0$ we shall consider as the first principal component.

Applying this algorithm to the recovery error, we find the second principal component, etc. The detailed algorithm of calculation of the principal components looks as follows:

So, let there is $n$ a component $X^0, X^1, \ldots, X^{n-1}$. It is required to construct $m$ domains $Y^0, Y^1, \ldots, Y^{m-1}$ so that recovery on these domains gave the least root-mean-square error of recovery on $m$ domains.

Thus, it is necessary to find a minimum of size

$$\sum_{k=0}^{n-1} \left\| X^k - \sum_{\mu=0}^{m-1} \alpha_k^\mu(m) Y^\mu \right\|_2 \qquad (7)$$

on all sets $Y^\mu$ and vectors $\alpha_k^\mu(m)$ such, that

$$\sum_{k=0}^{n-1} (\alpha_k^\mu(m))^2 = 1.$$

The numbers $\alpha_k^\mu(m)$ received as a result of the decision of this problem from $m$ do not depend, that, finally, allows to reduce to $m$ problems

$$\min\left\{\sum_{k=0}^{n-1}\left\|X^k - \alpha_k^\mu Y^\mu\right\|_2 \middle| Y^\mu, \alpha_k^\mu : \sum_{k=0}^{n-1}(\alpha_k^\mu)^2 = 1\right\} \qquad (8)$$

Let's consider the iterative process leading the decision of a problem (8), and at the same time and problems (7).

Let, in the beginning, $i = 0$ and $\alpha_k(0) = \frac{1}{\sqrt{n}}$. We shall calculate

331

$$Y(i) = \sum_{k=0}^{n-1} \alpha_k(i) X^k \qquad (9)$$

Further we shall calculate numbers $\alpha_k^*(i) = \left\langle Y(i), X^k \right\rangle$ and do normalization, that is

$$\alpha_k(i+1) = \frac{\alpha_k^*(i)}{\sqrt{\sum_{k=0}^{n-1}(\alpha_k^*(i))^2}}.$$

Let $i := i+1$ and we shall continue iterative process $N$ times where $N$ it is those that stabilizes either the domain $Y(i)$ and values $\alpha_k(i)$. As a rule, for this purpose it is enough to use 10-20 iterations. After that we believe $Y^0 = Y(N)$ and $\overline{\alpha}_k^0 = \alpha_k(N)$. Clearly, that the error of recovery of everyone components will be equal

$$\Delta X^k = X^k - \widetilde{X}^k,$$

where $\widetilde{X}^k = \overline{\alpha}_k^0 Y^0$ recovering of $k$ component on the domain $Y^0$.

The received error of recovery we shall perceive as a component to which (as initial data) we shall repeat the same iterative process, that is, we believe $i = 0$ and $\alpha_k(0) = \frac{1}{\sqrt{n}}$. It is calculated $Y(i) = \sum_{k=0}^{n-1} \alpha_k(i) \Delta X^k$ also numbers

$$\alpha_k^*(i) = \left\langle Y(i), \Delta X^k \right\rangle.$$

After normalization we get

$$\alpha_k(i+1) = \frac{\alpha_k^*(i)}{\sqrt{\sum_{k=1}^{n-1}(\alpha_k^*(i))^2}}.$$

Believing $i := i+1$ we shall continue iterative process. After stabilization of iterative process we get $Y^1 = Y(i)$ and $\alpha_k^1 = \alpha_k(i)$. Further we find $\Delta \widetilde{X}^k = \alpha_k^1 Y^1$.

And to the received error of recovery $\Delta^2 X^k = \Delta X^k - \Delta \widetilde{X}^k$ we shall iteratively apply the same process $n$ times.

At sufficient number of iterations $\Delta^n X^k = 0$ for all $k$, that is the algorithm realizes full decomposition a component $X^k$ on domains $Y^k (k = 0,1,\ldots,n-1)$.

Recovery on $m$ will be equal to domains

$$X_m^v = \sum_{k=0}^{m-1} \alpha_v^k Y^k .$$

Let's put the basic properties of the principal components

1. Equality takes place

$$\min\left\{\sum_{k=0}^{n-1}\left\|X^k - \sum_{\mu=0}^{m-1}\alpha_k^\mu(m)Y^\mu\right\|_2 \middle| Y^\mu, \alpha_k^\mu(m) : \sum_{k=0}^{n-1}(\alpha_k^\mu(m))^2 = 1\right\} =$$

$$= \sum_{k=0}^{n-1}\left\|X^k - \sum_{\mu=0}^{m-1}\alpha_k^\mu Y^\mu\right\|_2 .$$

2. If

$$\widehat{X}^v = \sum_{k=0}^{n-1} \alpha_v^k Y^k ,$$

that we have the Parseval equality

$$\sum_{v=0}^{n-1}\left\|\widehat{X}^v\right\|_2^2 = \sum_{v=0}^{n-1}\left\|Y^v\right\|_2^2 ,$$

3. Moreover, for $m = 1,\ldots,n$ and

$$\widehat{X}_m^v = \sum_{k=0}^{m-1}\alpha_v^k Y^k$$

that we have equation

$$\sum_{v=0}^{m}\left\|\widehat{X}_m^v - \widehat{X}^v\right\|_2^2 = \sum_{v=m+1}^{n-1}\left\|Y^v\right\|_2^2 .$$

4. Vectors $\alpha_0, \alpha_1, \ldots, \alpha_{n-1}$ form orthonormalized system.

5. Thus $\|Y^v\|_2^2 (v = 0, \ldots, n-1)$ is eigenvalue of the covariance matrix $\left[\langle X^i X^j \rangle\right]$, vectors $\alpha_v (v = 0, \ldots, n-1)$ its eigenvector.

6. Last frequency domains not structured data contain "white noise", that is.

## 4 Application of iterative algorithm of Karhunen-Loève Decomposition to construction of the text classifier

So, let are given $n$ vectors $X^0, X^1, \ldots, X^{n-1}$ and $m$ the principal components $Y^0, Y^1, \ldots, Y^{m-1}$ together with vectors of the form: $\alpha_k^\mu (m) k = 0, \ldots, m-1, \mu = 0, \ldots, n-1$.

Let's calculate distance between vector $Z$ and set $\{X^i\}_{i=0}^{n-1}$. According to Mahalanobis metrics

$$\left\|Z - \{X^i\}_{i=0}^{n-1}\right\|_{\Sigma^{-1}} = \sqrt{\left(Z - E\left(\{X^i\}_{i=0}^{n-1}\right)\right)\Sigma^{-1}\left(Z - E\left(\{X^i\}_{i=0}^{n-1}\right)\right)^T},$$

where $\Sigma$ covariance matrix.

Noticing, that for the principal components the covariance matrix is those, that on its diagonal coefficients is $\|Y^v\|_2^2 (v = 0, \ldots, m-1)$ and the others are equal to zero, we have

$$\Sigma^{-1} = \begin{pmatrix} \dfrac{1}{\|Y^0\|_2^2} & 0 & \ldots & 0 \\ 0 & \dfrac{1}{\|Y^1\|_2^2} & \ldots & 0 \\ \ldots & \ldots & \ldots & \ldots \\ 0 & 0 & \ldots & \dfrac{1}{\|Y^m\|_2^2} \end{pmatrix}$$

Besides if
$$\widehat{X} = \sum_{v=0}^{n-1} \widehat{X}^v = \sum_{v=0}^{n-1}\sum_{k=0}^{m-1} \alpha_v^k Y^k = \sum_{k=0}^{m-1}\sum_{v=0}^{n-1} \alpha_v^k Y^k = \sum_{k=0}^{m-1} \Lambda^k Y^k,$$

where
$$\Lambda^k = \sum_{v=0}^{n-1} \alpha_v^k,$$

Then ort of the central vector of set will be equal to
$$E\left(\{X^i\}_{i=0}^{n-1}\right) = \frac{\widehat{X}}{\|\widehat{X}\|_2}.$$

In order to calculate of distance it is necessary to receive decomposition of a checked vector on basis of mainstreams

$$\widehat{Z} = \sum_{k=0}^{m-1} \beta^k Y^k, \text{ where } \beta^k = \frac{\langle Z, Y^k \rangle}{\langle Y^k, Y^k \rangle}.$$

As a result can change normalization, therefore, it is necessary to normalize by unit.

Then the formula for calculation of Mahalanobis distance will look like

$$\left\|Z - \{X^i\}_{i=0}^{n-1}\right\|_{\Sigma^{-1}}^2 = \sum_{k=0}^{m-1} \frac{1}{\|Y^k\|_2^2}\left(\frac{\beta^k Y^k}{\left\|\sum_{k=0}^{m-1}\beta^k Y^k\right\|_2} - \frac{\Lambda^k Y^k}{\left\|\sum_{k=0}^{m-1}\Lambda^k Y^k\right\|_2}\right)^2 =$$

$$= \sum_{k=0}^{m-1}\left(\frac{\beta^k}{\left\|\sum_{k=0}^{m-1}\beta^k Y^k\right\|_2} - \frac{\Lambda^k}{\left\|\sum_{k=0}^{m-1}\Lambda^k Y^k\right\|_2}\right)^2$$

## 5 Reduction of dimension of classes

Capacious algorithms of calculation of the principal components essentially depends on the size of classes of initial data, therefore reduction of dimension of these classes is an actual problem. Further our reasoning are devoted to this problem. For simplicity of a statement as criterion of quality the size of scalar product will serve, thus, the class of individual vectors

335

(documents) is limited on sphere by a circle with the center in the end of the central vector of a class. At use of Mahalanobis distance of a point on the individual sphere, corresponding documents of one class, will be limited by curves of the second order.

For construction of a statistics files of the wordforms $b^\nu, \nu = 0,..., M-1$ belonging one class $B = \{b^\nu\}_{\nu=0}^{M-1}$ are consistently processed all. On set of wordforms of each processed text $b^\nu$ the set of unique (not repeating) wordforms and their counters is under construction- $(w_i^\nu, n_i^\nu)(i = 0,..., N^\nu - 1)$. Here $N^\nu$ - quantity of unique wordforms for the text $b^\nu$. After that data for each file are separately normalized

$$\overline{n}_i^\nu = \frac{n_i^\nu}{\sqrt{\sum_{j=0}^{N^\nu - 1}(n_j^\nu)^2}} (i = 0,..., N^\nu - 1).$$

After that, we order all words for each document in the same order (the word order is not essential, the main thing that words in each of structures $(w_i^\nu, n_i^\nu)(i = 0,..., N^\nu - 1)$ went in the same order) and we find the sum of all vectors $n_i(B) = \sum_{j=0}^{M-1} \overline{n}_i^\nu (i = 0,..., N(B))$ (where $N(B)$ - quantity of unique word forms for class $B$ as a whole) and it is normalized by its unit

$$\overline{n}_i(B) = \frac{n_i(B)}{\sqrt{\sum_{j=0}^{N(B)}(n_j(B))^2}}.$$

For the received central point of a class we form a file of statistics, writing down in it values $(w_i(B), \overline{n}_i(B))(i = 0,..., N(B))$.

For construction of the central vector of classes $\{B^\mu\}_{\mu=0}^{K-1}$ where each class $B^\mu$ is described by the central vector $(w_i(B^\mu), \overline{n}_i(B^\mu))(i = 0,..., N(B^\mu))$ it is necessary to find their sut, by summarizing all coordinates from all summable vectors for everyone values of a word form, that is for a word form $\omega$ we receive coordinate

$$n(\omega) = \sum_{\mu=0}^{K-1} \left\{ \overline{n}_i(B^\mu) \middle| w_i(B^\mu) = \omega, i = 0,...N(B^\mu) \right\}.$$

That is, it is necessary to make the list of unique word forms on all central vectors of classes $\{B^\mu\}_{\mu=0}^{K-1}$ and summarize their coordinates. As result will be the set, consisting their unique (not repeating) word forms and their coordinates

$$\left(w_i\left(\{B^\mu\}_{\mu=0}^{K-1}\right), n_i\left(\{B^\mu\}_{\mu=0}^{K-1}\right)\middle| i=0,...,N\left(\{B^\mu\}_{\mu=0}^{K-1}\right)\right)$$

where $N\left(\{B^\mu\}_{\mu=0}^{K-1}\right)$ is number of unique wordforms of set of classes $\{B^\mu\}_{\mu=0}^{K-1}$. It is necessary to normalize the received coordinates

$$\overline{n}_i\left(\{B^\mu\}_{\mu=0}^{K-1}\right) = \frac{n_i\left(\{B^\mu\}_{\mu=0}^{K-1}\right)}{\sqrt{\sum_{j=0}^{N\left(\{B^\mu\}_{\mu=0}^{K-1}\right)}\left(n_j\left(\{B^\mu\}_{\mu=0}^{K-1}\right)\right)^2}}$$

and the received vector $\left(w_i\left(\{B^\mu\}_{\mu=0}^{K-1}\right), \hat{n}_i\left(\{B^\mu\}_{\mu=0}^{K-1}\right)\middle| i=0,...,N\left(\{B^\mu\}_{\mu=0}^{K-1}\right)\right)$ will be the central vector of set $\{B^\mu\}_{\mu=0}^{K-1}$.

Ideally generated classification of a vector method is such set of classes $\{B^\mu\}_{\mu=0}^{K-1}$ for which the following condition $\forall b \in B^\mu, \mu = 0,...,K-1$ is satisfied takes place a parity

$$\langle \overline{n}(b), \overline{n}(B^\mu) \rangle < \langle \overline{n}(b), \overline{n}(B^\nu) \rangle, \nu \neq \mu. \qquad (10)$$

Let's notice, that application as criterion of quality of Mahalanobis distance will raise efficiency of algorithm, but will essentially increase resource capacity.

Let's consider a vector $\Lambda$ (control) of dimension $N(B^\mu)$ which coordinates accept only one of two admissible values

$$\lambda_i = \begin{cases} 0, \\ 1. \end{cases}$$

Through $\Lambda b$ we shall designate direct product of vectors $\Lambda$ and $b$, that is

$$\Lambda b = \left(\lambda_0 \overline{n}_0(b), \lambda_1 \overline{n}_1(b), ..., \lambda_{N(B^\mu)} \overline{n}_{N(B^\mu)}(b)\right).$$

Control $\Lambda$ we shall name allowed on a class $B^\mu = \{b^k\}_{k=0}^{M-1}$ if the condition

$$\langle \overline{\Lambda \overline{n}}(b^k), \overline{\Lambda \overline{n}}(B^\mu) \rangle < \langle \overline{\Lambda \overline{n}}(b^k), \overline{n}(B^\nu) \rangle, \nu \neq \mu, k = 0,1,...,M-1 \qquad (11)$$

is satisfied.

Allowed control such that $\sum_{k=0}^{M-1}(\Lambda b^k)^2 \to \max$ is named optimum.

If for $\nu \neq \mu$ set of allowed control is singular, the class $B^\mu = \{b^k\}_{k=0}^{M-1}$ is certain incorrectly, that is it is inseparable with a class $B^\nu$.

The problem of a finding of optimum control of classical methods is complex enough, therefore we shall apply genetic algorithms to its decision. Main principles of work in genetic algorithms are concluded in the following scheme ([4]):

1. We generate an initial population from n chromosomes $\lambda_i$.
2. We calculate for each chromosome its suitability that is performance of a condition (2).
3. We choose pair chromosomes-parents by means of one of ways of selection.
4. We generate posterity of the chosen parents, using genetic operators, first of all crossover and mutation.
5. We repeat steps 3–4, the new generation of a population containing n of chromosomes will not be generated yet.
6. We repeat steps 2–5, the criterion of the termination of process will not be reached yet.

As criterion of the termination of process the set quantity of generations or a convergence of a population can serve.

There are some approaches to a choice of parental pair. Selection consists that parents can become only those individuals which value of fitness not less than threshold size, in our case, average value of fitness on a population. Such approach provides faster convergence of algorithm.

The recombination operator is applied directly after the operator of selection of parents to reception of new individuals-descendants. The sense of recombination consists that the created descendants should inherit the genetic information from both parents. Recombination of binary lines it is accepted to name crossover. In our case it is used single-point crossover which is modeling as follows-let there are two parental individuals with chromosomes $X = \{x_i, i \in \{0,...,L\}\}$ and $Y = \{y_i, i \in \{0,...,L\}\}$.

We randomly define a point inside of a chromosome (a point of break) in which both chromosomes share on two parts and exchange them. After process of reproduction there are mutations (mutation). The given operator is necessary for «knocking-out» a population from a local extremum and interferes with premature convergence. It is reached because casually chosen gene in a chromosome changes.

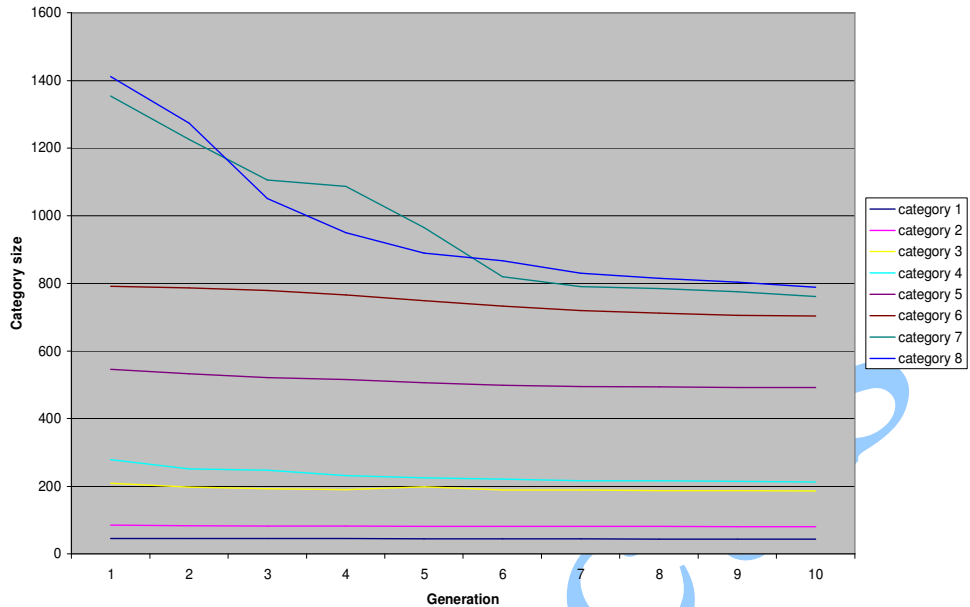

Figure 1 - the diagram of categories dimension reduction at use of genetic algorithms

For creation of a new population it is possible to use various methods of selection of individuals. We use elite selection. The intermediate population which includes both parents, and their descendants is created. Members of this population are estimated, and behind that of them get out N the best (suitable) which will enter into following generation.

The result of application of genetic algorithm to a problem of reduction of dimension of a class, is resulted in figure 1.

The combination of use of an iterative method of construction of Mahalanobis distance together with the offered design of reduction of dimension of classes allows us to receive effective algorithms of classification for greater bases of documents.

## 6 Conclusions

Results of work have indicate that, for reduction of dimension of vectors of classes of documents effectively enough to use genetic algorithms with chromosomes in length equal to quantity of nonzero coordinates of the central vector and binary genes. It is offered to spend minimization of a genotype from a condition of a minimality of influence on a class at maintenance of the set degree of localization of a class. For test base "Reuters" with condition of hit in a class not less than 90 % of documents, it was possible to reduce dimension of classes from 10 % to 50 %.

**References**

[Bac96]  Back Thomas - *Evolutionary Algorithms in Theory and Practice,* Oxford University Press, New York, 1996

[BB06]   Berry Michael W., Browne Murray - *Lecture notes in data mining,* World Scientific Publishing Co, Pte, Ltd.: Singapore, 2006

[GKWZ07] Gorban A. N., Kegl B., Wunsch D., Zinovyev A. Y. (Eds.), *Principal Manifolds for Data Visualization and Dimension Reduction*, Series: Lecture Notes in Computational Science and Engineering 58, Springer, Berlin - Heidelberg - New York, 2007

[SSL08]  Shumeyko A.A., Sotnik S.L., Lysak M.V. - *Using genetic algorithms in texts classification problems,* In VI International Conference "Mathematical methods and Programming in Intelligence Systems" (MPZIS-2008), Dnipropetrovsk, 2008

[TG74]   Tou J., Gonzalez R. - *Recognition principles*, Addison-Wesley Publishing Company, 1974